\documentclass{article}

\usepackage[final]{nips_2016}

\usepackage[utf8]{inputenc} 
\usepackage[T1]{fontenc}    
\usepackage{url}            
\usepackage{booktabs}       
\usepackage{amsfonts}       
\usepackage{nicefrac}       
\usepackage{microtype}      
\usepackage{float}
\usepackage{amsmath}

\usepackage{graphicx}
\usepackage{algorithm}
\usepackage{algorithmic}
\usepackage{enumitem}

\title{Interpreting Finite Automata for Sequential Data}

\author{
  Christian Albert Hammerschmidt\\
  SnT\\
University of Luxembourg\\
\texttt{christian.hammerschmidt@uni.lu} \\
  \And
  Sicco Verwer \\
  Department of Intelligent Systems\\
  Delft University of Technology\\
  \texttt{s.e.verwer@tudelft.nl} \\
  \AND
  Qin Lin \\
  Department of Intelligent Systems\\
  Delft University of Technology\\
  \texttt{q.lin@tudelft.nl}
  \And
  Radu State \\
  SnT\\
  University of Luxembourg \\
  \texttt{radu.state@uni.lu} \\
}

\newcommand{\parencite}[1]{\citep{#1}}
\newcommand{\textcite}[1]{\citet{#1}}

\begin{document}

\maketitle

\begin{abstract}
Automaton models are often seen as \emph{interpretable} models. Interpretability itself is not well defined: it remains unclear what interpretability means without first explicitly specifying objectives or desired attributes. In this paper, we identify the key properties used to interpret automata and propose a modification of a state-merging approach to learn variants of finite state automata. We apply the approach to problems beyond typical grammar inference tasks. Additionally, we cover several  use-cases for prediction, classification, and clustering on sequential data in both supervised and unsupervised scenarios to show how the identified key properties are applicable in a wide range of contexts. 
\end{abstract}
\vspace{0.35cm}
\section{Introduction}
The demand for explainable machine learning is increasing, driven by the spread of machine learning techniques to sensitive domains like cyber-security, medicine, and smart infrastructure among others. Often, the need is abstract but nevertheless a requirement, e.g., in the recent EU regulation \parencite{goodman_european_2016}.\newline
Approaches to explanations range from post-hoc explanation systems like \textcite{turner_model_2015}, which provide explanations of decisions taken by black-box systems to the use of linear or specialized white-box systems \parencite{fiterau_trade-offs_2012} that generate models seen as simple enough for non-expert humans to interpret and understand. \textcite{lipton_mythos_2016} outlines how different motivations and requirements for interpretations lead different notions of interpretable models in supervised learning. \newline
Automata models, such as (probabilistic) deterministic finite state automata ((P)DFA) and timed automata (RA) have long been studied (\textcite{hopcroft_introduction_2013})
and are often seen as interpretable models. Moreover, they are learnable from data samples, both in supervised and unsupervised (see~\textcite{higuera_grammatical_2010}) fashion. But which properties make these models interpretable, and how can we get the most benefit from them? We argue that---especially in the case of unsupervised learning---automata models have a number of properties that make it easy for humans to understand the learned model and project knowledge into it: a graphical representation, transparent computation, generative nature, and our good understanding of their theory.
%
%
\section{Preliminaries}
\textbf{Finite automata.} The models we consider are variants of deterministic finite state automata, or finite state machines. These have long been key models for the design and analysis of computer systems~\parencite{lee_principles_1996}. We provide a conceptual introduction here, and refer to Section \ref{apx:sec:finite} in the appendix for details. An automaton consists of a set of states, connected by transitions labeled over an alphabet.
It is said to accept a word (string) over the alphabet in a computation if there exists a path of transitions from a predefined start state to one of the predefined final states, using transitions labeled with the letters of the word. Automata are called deterministic when there exists exactly one such path for every possible string. Probabilistic automata include probability values on transitions and compute word probabilities using the product of these values along a path, similar to hidden Markov models (HMMs).

\textbf{Learning approaches.} As learning finite state machines has long been of interest in the field of grammar induction, different approaches ranging from active learning algorithms \parencite{angluin_learning_1987} to algorithms based on the method of moments \parencite{balle_spectral_2014} have been proposed. Process mining~\parencite{van_der_aalst_process_2012} can also be seen as a type of automaton learning, focusing on systems that display a large amount of concurrency, such as business processes, represented as interpretable Petri-nets. We are particularly interested in state-merging approaches, based on \textcite{oncina_identifying_1992}. While Section \ref{apx:sex:state-merging} in the appendix provides formal details, we provide a conceptual introduction here.\newline
The starting point for state-merging algorithms is the construction of a tree-shaped automaton from the input sample, called augmented prefix tree acceptor (APTA). It contains all sequences from the input sample, with each sequence element as a directed labeled transition. Two samples share a path if they share a prefix. The state-merging algorithm reduces the size of the automaton iteratively by reducing the tree through merging pairs of states in the model, and forcing the result to be deterministic. The choice of the pairs, and the evaluation of a merge is made heuristically: Each possible merge is evaluated and scored, and the highest scoring merge is executed. This process is repeated and stops if no merges with high scores are possible. These merges generalize the model beyond the samples from the training set: the starting prefix tree is already an a-cyclic finite automaton. It has a finite set of computations, accepting all words from the training set. Merges can make the resulting automaton cyclic. Automata with cycles accept an infinite set of words. State-merging algorithms generalize by identifying repetitive patterns in the input sample and creating appropriate cycles via merges. 
Intuitively, the heuristic tries to accomplish this generalization by identifying pairs of states that have similar future behaviors. In a probabilistic setting, this  similarity might be measured by similarity of the empirical probability distributions over the outgoing transition labels. In grammar inference, heuristics rely on occurrence information and the identity of symbols, or use global model selection criteria to calculate merge scores. 
%
\section{Flexible State-Merging}
\label{sec:flexible}
The key step in state-merging algorithms is the identification of good merges. A merge of two states is considered good if the possible futures of the two states are very similar.
By focusing on application-driven notions of \emph{similarity} of sequences and sequence elements, we modify the state-merging algorithm as follows: For each possible merge, a heuristic can evaluate an application-driven similarity measure to obtain the merge score. Optionally,
each symbol of the words is enriched with addition data, e.g. real values. 
This information can be aggregated in each state, e.g. by averaging. It is used to guide the heuristic in reasoning about the similarity of transitions and therefore the inferred latent states, or guide a model selection criterion, e.g. by using mean-squared-error minimization as an objective function.
It effectively separates the symbolic description connecting the latent variables from the objective (function) used to reason about the similarity. An implementation in C++ is available\footnote{\url{https://bitbucket.org/chrshmmmr/dfasat}}.
The importance of combining data with symbolic data is getting renewed attention, c.f. recent works such as \textcite{garnelo_towards_2016}.
In Section \ref{sec:application}, we outline regression automata as a use case of our approach.
%
\section{Aspects of Interpretability in Automata}
\label{sec:aspects}
To understand how a particular model works as well as how to go beyond the scope of the model and combing foreground knowledge about the application with the model itself, we now focus on which aspects of automata enable interpretations: 
\begin{enumerate}
  \setlength\itemsep{0.25em}
\item Automata have an easy \emph
{graphical representation} as cyclic, directed, labeled graphs, offering a hierarchical view of sequential data. 
\item \emph{Computation is transparent}. Each step of the computation is the same for each symbol $w_i$ of an input sample $w$. It can be verified manually (e.g. visually), and compared to other computation paths through the latent state space. This makes it possible to analyze training samples and their contribution to the final model.
\item Automata are \emph{generative}. Sampling from the model helps to understand what it describes. Tools like model checkers to query properties in a formal way, e.g. using temporal logic, 
can help to analyze the properties of the model.
\item Automata are \emph{well studied} in theory and practice, including composition and closure properties, sub-classes and related equally expressive formalisms. 
This makes it easy for humans to transfer their knowledge onto it: The model is frequently used in system design as a way to describe system logic, and are accessible to a wide audience.
\end{enumerate}
The intention behind using either of these aspects depends on the purpose of the interpretation, e.g. trust, transparency, or generalizing beyond the input sample. Especially for unsupervised learning, we believe that knowledge transfer and exploratory knowledge discovery are common motivations, e.g. in (software) process discovery.
%
\vspace{-0.35cm}
\section{Application Case Studies}
\label{sec:application}
In the following we present some use cases of automata models and how they are interpreted and how the properties identified in Section \ref{sec:aspects} contribute to it. While this is by no means an exhaustive literature study, we hope that it helps to illustrate how the different aspects of interpretability are used in practice.
In unsupervised learning, the data is observations without labels or counter-examples to the observed events. Often, there is no ground-truth to be used in an objective function. Methods for learning such systems typically use statistical assumptions to compute state similarity.

\textbf{Software systems.} Automata models are often used to infer models of software 
in an unsupervised fashion, e.g. \textcite{walkinshaw_reverse_2007}. In these cases, the generative property of automaton models is see as interpretable: It is possible to ask queries using a temporal logic like LTL \parencite{clarke_temporal_2005} to answer questions regarding the model, e.g. whether a certain condition will eventually be true, or analyze at what point a computation path deviated from the expected outcome. In \textcite{smetsers_complementing_2016}, the authors use this property to test and validate properties of code by first fuzzing the code to obtain execution paths and the associated inputs and then learn a model to check LTL queries on.\newline
Additionally, we can transfer human expert knowledge on system design \parencite{wagner_modeling_2006} to inspect the model, e.g. to identify the function of substructures identified. An example can be found in \textcite{smeenk_applying_2015}, where the authors infer a state machine for a printer via active learning. Through visual inspection alone it is possible to identify deadlocks that are otherwise hard to see in the raw data. The visual analysis helped to identify bugs in the software the developers were unaware of. The appendix shows the final model in Figure \ref{fig:printer}.

\textbf{Biological systems.} In \textcite{schmidt_online_2014}, timed automata are used to infer the cell cycle of yeast based on sequences of gene expression activity. The graphical model obtained (c.f. Figure \ref{fig:jana}) can be visually compared to existing models derived using other methods and combined with a-priori knowledge in biology.

\textbf{Driver behavior.} In our ongoing work using the RTI+ state-merging algorithm for timed automata \parencite{verwer_efficiently_2008},  we analyze car following behavior of human drivers. The task is to relate driver actions like changes of vehicle speed (and thus distance and relative speed to a lead vehicle) to a response action, e.g. acceleration. The inferred automaton model is inspected visually like a software controller. Different driving behaviors are distinguished by clustering the states of the automaton, i.e. the latent state space. The discovered distinct behaviors form control loops within the model. Figure \ref{fig:qin} in the appendix shows an example with the discovered clusters highlighted.
%

\textbf{Wind speed prediction.} In our own work, we applied automata learning in different ways to a problem not related to grammar inference, predicting short-term changes in wind speeds. We take two different approaches to obtain models that tell us more about the data than just a minimizer of the objective function: In one approach, \parencite{pellegrino_learning_2016}, we \textbf{discover structure} in the data by using 
by inferring transition guards over a potentially infinite alphabet, effectively discovering a clustering as transition labels from the sequences automatically. The only constraint and objective used here is the similarity of future behaviors. The learned model can be seen as a structure like a decision tree built in a bottom-up fashion, but allowing loops for repetitive patterns. Figure \ref{fig:nino_rag} in the appendix shows an example of such an automaton. In another approach \parencite{qin_short-term_2016}, we use our flexible state-merging framework to \textbf{impose structure} through parameter choices. We derive discrete events from the physical wind speed observations by using clustering approaches to obtain a small alphabet of discrete events as a symbolic representation that fits the underlying data well. Using a heuristic that scores merges based on minimizing a mean squared error measure, the automata model has a good objective function for regression, as well as discovers latent variables in terms of the given discretization.
In practice, other choices of discretization can be used. By using thresholds of the turbine used in wind mills, e.g. the activation point, one could infer a model whose latent states relate to the physical limitations of the application. We are planning to analyze this approach in future work. As with the previous example, the learned model can be seen as a description of decision paths taken in the latent state-space. If the model makes predictions that are difficult to believe for human experts, the computation and the model prediction can be visually analyzed to see which factors contributed to it, and how the situation relates to similar sets of features. %
\vspace{-0.35cm}
\section{Discussion}
\vspace{-0.15cm}
The applications surveyed in Section \ref{sec:application} show that \emph{interpreting} finite state automata as models takes many forms and serves different goals. As such, this interpretation is not a feature inherent in the models or the algorithms themselves. Rather, interpretations are defined by the need and intention of the user. But yet, \textbf{interpretations of automata models draw from a core set of properties} as identified in Section \ref{sec:aspects}:  graphical representations, transparent computation, generative nature, and our understanding of their theory.
\newline
We note that \textbf{automata models are particularly useful in unsupervised learning:}
Applications of automata models often aim at easing the transfer of knowledge about the subject, or related subjects, to the data generating process. In this case, machine learning serves as a tool for exploration, to deal with epistemic uncertainty in observed systems. The goal is not only to obtain a more compact view of the data, but learn how to generalize from the observed data. Often, it is often unclear what the a-priori knowledge is as users rely on experience. This makes it very difficult to formalize a clear objective function. A visual model with a traceable computation helps to guide the users, and helps to iterate over multiple models.\newline
\textbf{Flexible state-merging allows to obtain automata models in new domains:} in our flexible state-merging framework presented in Section \ref{sec:flexible}, we try to separate the symbolic representation from the objective function and heuristic. We hope that this will help to guide discovery by stating the model parameters, e.g. the symbols, independently form the heuristic that guides the discovery of latent variables. In this fashion, it is possible to learn models with interpretable aspects without having to sacrifice the model performance on the intended task.
\newline

\textbf{Future work.} We hope that this discussion will help to build a bridge between practitioners and experts in applied fields on one side, and the grammar inference and machine learning community on other side. As probabilistic deterministic automata models are almost as expressive as HMMs, the models and techniques described here are applicable to a wide range of problems with decent performance metrics.
We see a lot of promise in combining symbolic data with numeric or other data via a flexible state-merging approach to bring automata learning to fields beyond grammatical inference.

\subsubsection*{Acknowledgments}
I would like to thank, in no particular order, my colleagues, Joshua Moermann, Rick Smeters, Nino Pellegrino, Sara Messelaar, Corina Grigore, and Mijung Park for their time and feedback on this work. This work is partially funded by the FNR AFR grant PAULINE and Technologiestichting STW VENI project 13136 (MANTA) and NWO project 62001628 (LEMMA).

\newpage
\small
\setlength{\bibsep}{4pt plus 0ex}
\bibliographystyle{plainnat}
\bibliography{minimalbib}
\normalsize
\newpage

\appendix
\section{Visualizations of Automata}
\label{apx:vis}
\vspace{-0.75cm}
\begin{figure}[H]
\centering
\includegraphics[scale=0.25]{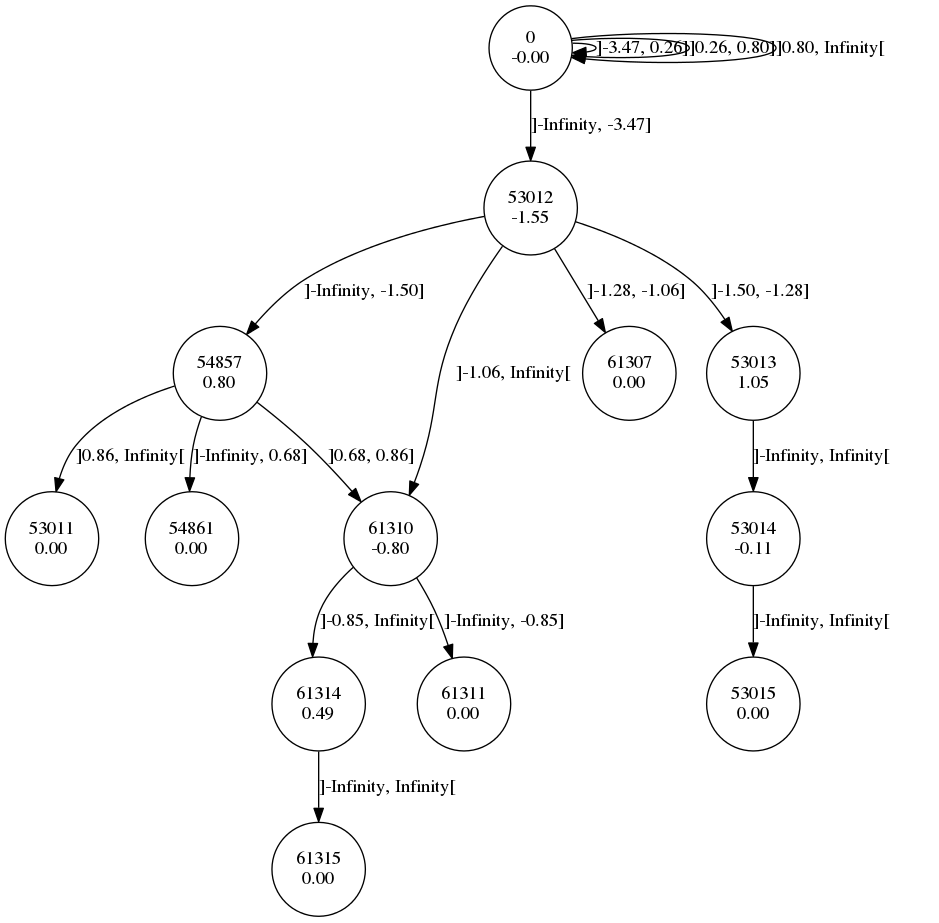}
\caption{Auto-regressive model of short-term average wind speeds using the approach presented by \textcite{qin_short-term_2016}. The ranges on the transitions indicate for which speed they activate, the number in the nodes is the predicted change in speed. The top node models persistent behavior (i.e. no change in speed for the next period), whereas the lower part denotes exceptions to this rule.}
\label{fig:nino_rag}
\end{figure}
\begin{figure}[H]
\centering
\includegraphics[scale=0.425,angle=0]{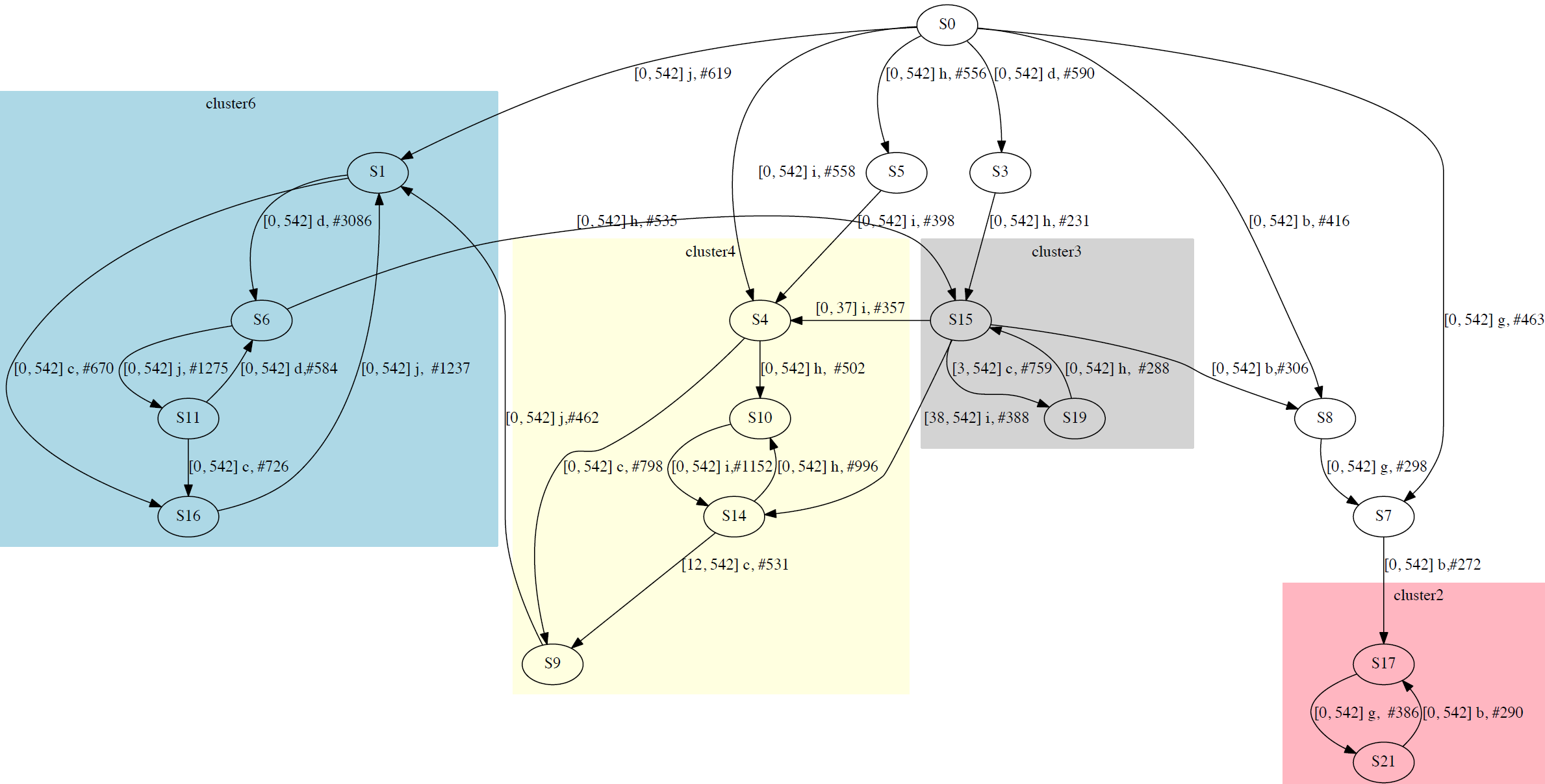}
\caption{Car-following controller model with highlighted clusters. Each color denotes one distinct cluster. The loop in cluster 6 (colored light blue), e.g. state sequence: 1-6-11-16-1 with symbolic transitions loop: d (keep short distance and slower than lead vehicle)-j (keep short distance and same speed to lead vehicle)-c (keep short distance and faster than lead vehicle))-j, can be interpreted as the \textbf{car-following behavior at short distances}, i.e. adapting the speed difference with the lead vehicle around 0 and bounding the relative distance in small zone. Similarly interesting and significant loops can be also seen in cluster 2 (colored pink) and cluster 4 (colored light yellow), which are \textbf{long distance} and \textbf{intermediate distance car-following} behaviors respectively. The intermediate state like $S15$ in cluster 3 (colored light grey) explains how to switch between clusters.}
\label{fig:qin}
\end{figure}
\begin{figure}[H]
\centering
\includegraphics[scale=0.65,angle=0]{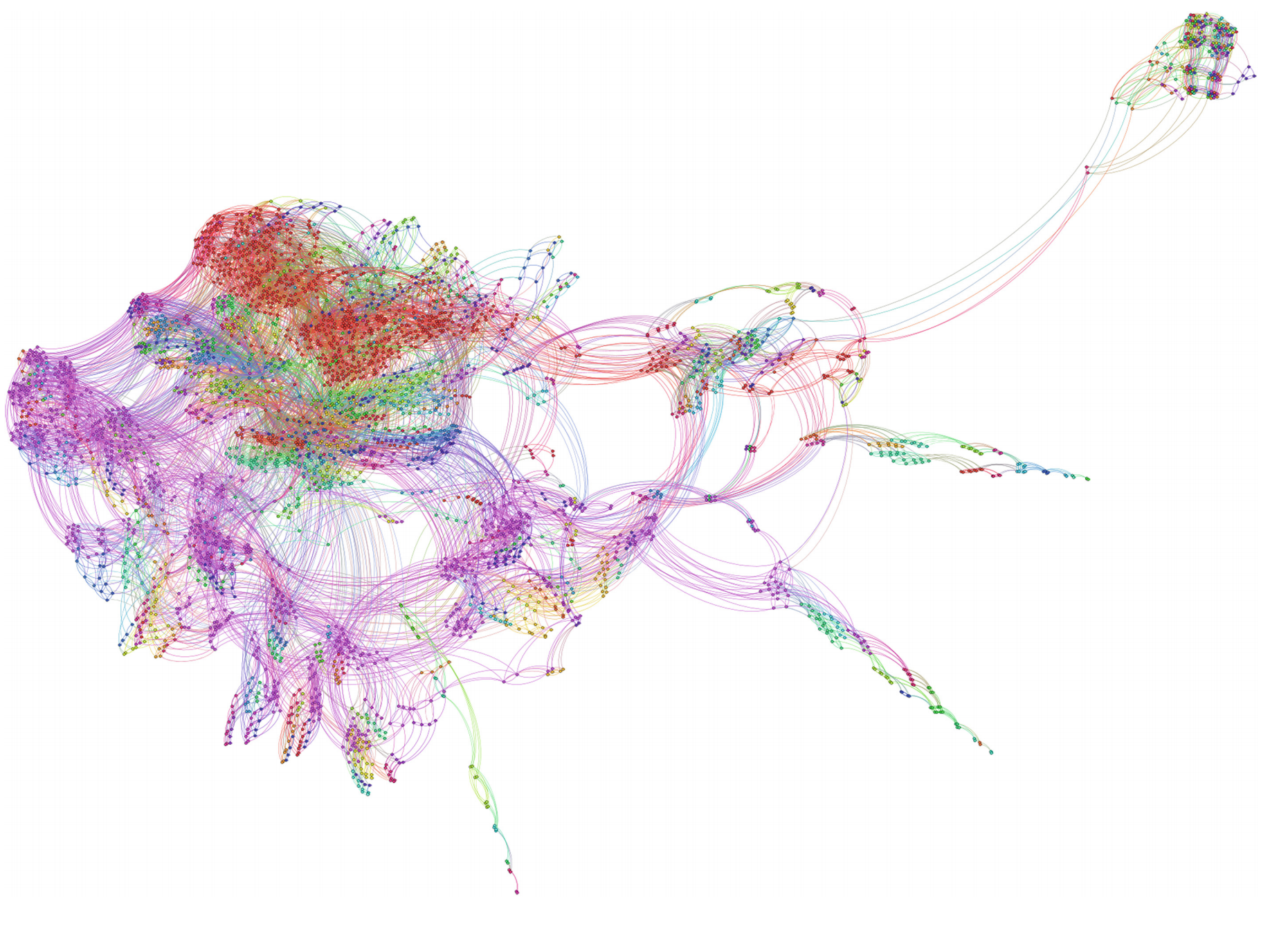}
\caption{Visualization of a state machine learned from a printer controller using an active learning approach. Each state, represented as a dot, represents a state in the controller software and each transition, represened as an edge, a state-change upon receiving input. Using knowledge about the controller design, it is possible to identify the sparsely connected protrusions of states at the bottom as deadlock situations. Taken from \textcite{smeenk_applying_2015}}.
\label{fig:printer}
\end{figure}
\begin{figure}[H]
\centering
\includegraphics[scale=0.66,angle=0]{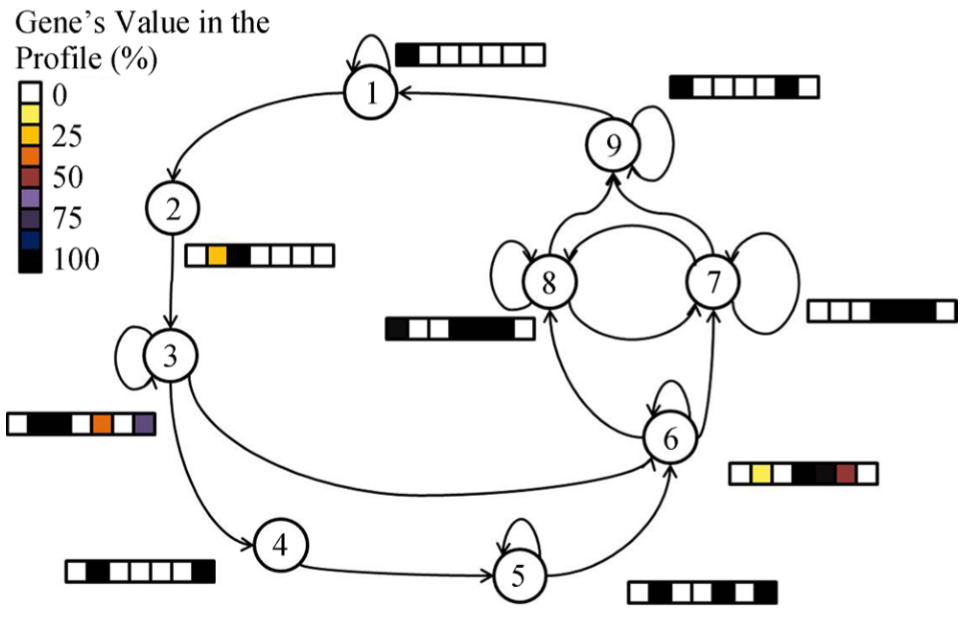}
\caption{The yeast cell cycle learned in \textcite{schmidt_online_2014} using a passive online learning algorithm for timed automata. The colored bars indicate gene activities. The model corresponds to the well-known yeast cycle.}
\label{fig:jana}
\end{figure}
\newpage
\section{Finite State Automata}
\label{apx:sec:finite}
\begin{figure}[H]
    \begin{minipage}{.6\textwidth}
\includegraphics[scale=0.8]{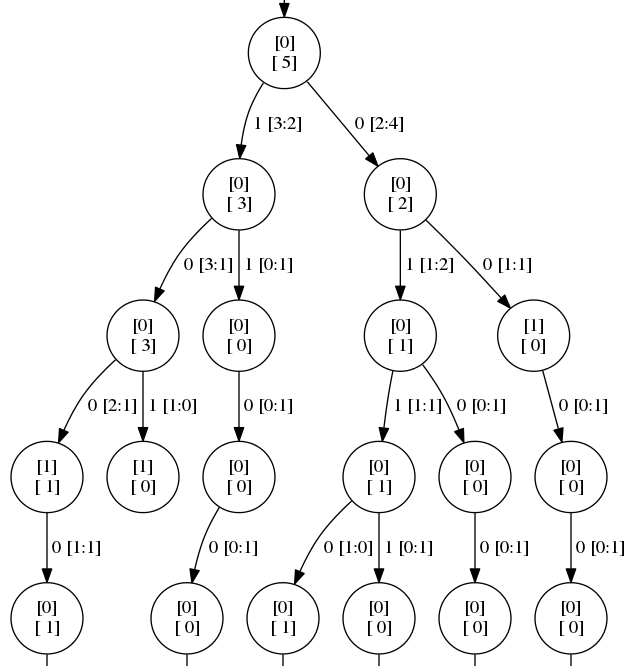}
        \caption{The initial part of the prefix tree, built from the input sample as shown in Figure \ref{fig:input}. The square brackets indicate occurrence counts of positive data. }
        \label{fig:prefix}
    \end{minipage}%
    \hspace{0.02\textwidth}
    \begin{minipage}{0.4\textwidth}
        \centering
        \includegraphics[scale=0.5]{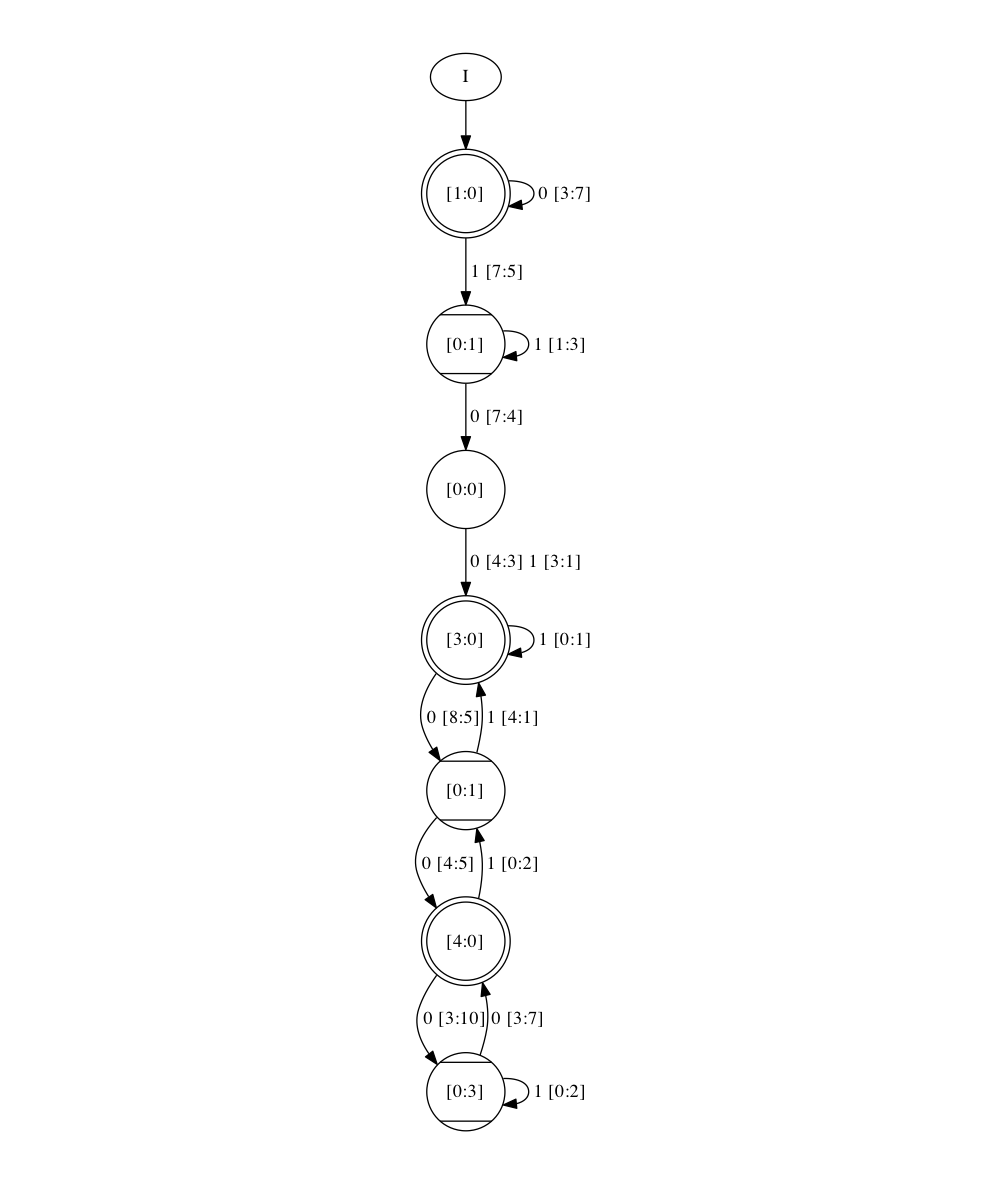}
        \caption{An automaton learned from the input sample. The numbers in square brackets indicate occurrence counts of positive and negative samples.}
        \label{fig:sample}
    \end{minipage}
\end{figure}

\normalsize
In this section, we give a formal introduction to deterministic finite state automata as the most basic model considered in the related work. For other variants like probabilistic automata, timed and real time automata, and regression automata, we refer to the cited papers for formal introductions.\newline
A \emph{deterministic finite state automaton} (DFA) is one of the basic and most commonly used finite state machines. Below, we provide a concise description of DFAs, the reader is referred to~\textcite{hopcroft_introduction_2013} for a more elaborate overview. A DFA $\mathit{A}=\left<Q,T,\Sigma,q_0,Q_+\right>$ is a directed graph consisting of a set of \emph{states} $Q$ (nodes) and labeled \emph{transitions} $T$ (directed edges). An example is shown in Figure~\ref{fig:sample}. The \emph{start state} $q_0 \in Q$ is a specific state of the DFA and any state can be an \emph{accepting state} (final state) in $Q_+ \subseteq Q$. The labels of transitions are all members of a given \emph{alphabet} $\Sigma$. A DFA $\mathit{A}$ can be used to \emph{generate} or \emph{accept} sequences of symbols (strings) using a process called \emph{DFA computation}. This process begins in $q_0$, and iteratively \emph{activates} (or \emph{fires}) an outgoing transition $t_i = \left< q_{i-1}, q_i, l_i \right> \in T$ with label $l_i \in \Sigma$ from the \emph{source state} it is in $q_{i-1}$, moving the process to the \emph{target state} $q_i$ pointed to by $t_i$. A computation $q_0t_1q_1t_2q_2\ldots t_nq_n$ is \emph{accepting} if the state it \emph{ends} in (its last state) is an accepting state $q_n \in Q_+$, otherwise it is \emph{rejecting}. The labels of the activated transitions form a string $l_1 \ldots l_n$. A DFA accepts exactly those strings formed by the labels of accepting computations, it rejects all others. Since a DFA is \emph{deterministic} there exists exactly one computation for every string, implying that for every state $q$ and every label $l$ there exists at most one outgoing transition from $q$ with label $l$. A string $s$ is said to \emph{reach} all the states contained in the computation that forms $s$, $s$ is said to \emph{end} in the last state $q_n$ of such a computation. The set of all strings accepted by a DFA $\mathit{A}$ is called the \emph{language} $L(\mathit{A})$ of $\mathit{A}$.

\section{State-Merging Algorithms}
\label{apx:sex:state-merging}
The idea of a state-merging algorithm is to first construct a tree-shaped DFA $\mathit{A}$ from the input sample $S$, and then to merge the states of $\mathit{A}$. This DFA $\mathit{A}$ is called an \emph{augmented prefix tree acceptor} (APTA). An example is shown in Figure~\ref{fig:prefix}. For every state $q$ of $\mathit{A}$, there exists exactly one computation that ends in $q$. This implies that the computations of two strings $s$ and $s'$ reach the same state $q$ if and only if $s$ and $s'$ share the same prefix until they reach $q$. 
Furthermore, an APTA $\mathit{A}$ is constructed to be \emph{consistent} with the input sample $S$, i.e., $S_+ \subseteq L(\mathit{A})$ and $S_- \cap L(\mathit{A}) = \emptyset$. Thus a state $q$ is accepting only if there exists a string $s \in S_+$ such that the computation of $s$ ends in $q$. Similarly, it is rejecting only if the computation of a string $s \in S_-$ ends in $q$. As a consequence, $\mathit{A}$ can contain states that are neither accepting nor rejecting. No computation of any string from $S$ ends in such a state. Therefore, the rejecting states are maintained in a separate set $Q_- \subseteq Q$, with $Q_- \cup Q_+ = \emptyset$. Whether a state $q \in Q \setminus (Q_+ \cup Q_-)$ should be accepting or rejecting is determined by merging the states of the APTA and trying to find a DFA that is as small as possible. 
\begin{figure}[H]
\centering
\begin{verbatim}

1 9 1 0 0 0 0 0 0 0 0
1 15 0 1 1 0 0 0 1 0 1 0 1 0 1 0 0
0 5 0 1 0 0 0
1 2 0 0
0 12 1 1 0 0 0 0 0 0 0 1 0 0
1 5 1 0 1 0 0
1 3 1 0 0
0 6 0 0 0 0 0 1
1 3 1 0 1
0 20 1 0 0 0 0 0 1 0 1 0 0 0 0 0 1 1 1 0 0 0
0 13 0 1 1 1 0 1 0 0 0 0 0 0 0
1 3 1 0 1
1 7 1 0 0 0 0 0 0
\end{verbatim}
\caption{Input sample for the APTA in Figure \ref{fig:prefix} and the learned model in Figure \ref{fig:sample}. The first column indicates whether the sample is positive or negative (i.e. a counter-example), the second column indicates the length of the word. The following symbols form the word, with each symbol separated by a space. This input format is commonly used in tools in the grammar induction community.}
\label{fig:input}
\end{figure}
A \emph{merge} 
of two states $q$ and $q'$ combines the states into one: it creates a new state $q''$ that has the incoming and outgoing transitions of both $q$ and $q'$, i.e., replace all $\left<q, q_t, l \right>, \left<q', q_t, l \right> \in T$ by $\left< q'', q_t, l \right>$ and all $\left<q_s, q, l \right>, \left<q_s, q', l \right> \in T$ by $\left< q_s, q'', l \right>$. Such a merge is only allowed if the states are \emph{consistent}, i.e., it is not the case that $q$ is accepting while $q'$ is rejecting or vice versa. When a merge introduces a non-deterministic choice, i.e., $q''$ is now the source of two transitions $\left<q'',q_1,l\right>$ and $\left<q'',q_2,l\right>$ in $T$ with the same label $l$, the target states of these transitions $q_1$ and $q_2$ are merged as well. This is called the \emph{determinization} process (c.f. the while-loop in Algorithm \ref{alg:merge}), and is continued until there are no non-deterministic choices left. However, if this process at some point merges two inconsistent states, the original states $q$ and $q'$ are also considered inconsistent and the merge will fail. The result of a successful merge is a new DFA that is smaller than before, and still consistent with the input sample $S$. A state-merging algorithm iteratively applies this state merging process until no more consistent merges are possible. The general algorithm is outlined in Algorithm \ref{alg:red_blue}. Figure \ref{fig:sample} shows an automaton obtained from the input given in Figure \ref{fig:input}, which is also depicted as an APTA in Figure \ref{fig:prefix}.
\begin{algorithm}[t]
\caption{State-merging in the red-blue framework\label{alg:red_blue}}
\begin{algorithmic}
\REQUIRE an input sample $S$
\ENSURE $\mathit{A}$ is a DFA that is consistent with $S$
\STATE $\mathit{A}$ = {\sf apta}$(S)$ \COMMENT{construct the APTA $\mathit{A}$}
\STATE $R = \{q_0\}$ \COMMENT{color the start state of $\mathit{A}$ red}
\STATE $B = \{q \in Q \setminus R \mid \exists \left<q_0, q, l\right> \in T\}$ \COMMENT{color all its children blue}
\WHILE[while $\mathit{A}$ contains blue states]{ $B \not= \emptyset$ }
\IF[if there is a blue state inconsistent with every red state]{ $\exists b \in B$ s.t. $\forall r \in R$ holds $merge(\mathit{A},r,b) =$ {\sc false}}
\STATE $R := R \cup \{b\}$ \hfill // color $b$ red
\STATE $B := B \cup \{q \in Q \setminus R \mid \exists \left<b, q, l\right> \in T\}$ \COMMENT{color all its children blue}
\ELSE
\FORALL[forall red-blue pair of states] {$b \in B$ and $r \in R$ }
\STATE compute the $\textsf{evidence}(\mathit{A},q,q')$ of $merge(\mathit{A},r,b)$ \COMMENT{find the best performing merge}
\ENDFOR
\STATE $\mathit{A} := merge(\mathit{A},r,b)$ with highest {\sf evidence} \COMMENT{perform the best merge}
\STATE let $q''$ be resulting state
\STATE $R := R \cup \{q''\}$ \COMMENT{color the resulting state red}
\STATE $R := R \setminus \{r\}$ \COMMENT{uncolor the merged red state}
\STATE $B := \{q \in Q \setminus R \mid \exists r \in R \text{ and } \left<r, q, l\right> \in T\}$ \COMMENT{recompute the set of blue states}
\ENDIF
\ENDWHILE
\RETURN $\mathit{A}$
\end{algorithmic}
\end{algorithm}
\begin{algorithm}
\caption{Merging two states: {\sf merge} ($A$, $q$, $q'$) \label{alg:merge}}
\begin{algorithmic}
\REQUIRE an augmented DFA $\mathit{A} = \left<Q,T,\Sigma,q_0,Q_+,Q_-\right>$ and two states $q,q' \in Q$
\ENSURE if $q$ and $q'$ are inconsistent, return {\sc false}; else return $A$ with $q$ and $q'$ merged.
\IF{ ($q \in Q_+$ and $q' \in Q_-$) or ($q \in Q_-$ and $q' \in Q_+$)}
\RETURN {\sc false} \COMMENT{return {\sc false} if $q$ is inconsistent with $q'$}
\ENDIF
\STATE let $A'\left<Q',T',\Sigma,q_0',Q_+',Q_-'\right>$ be a copy of $A$ \COMMENT{initialize the result $\mathit{A}'$}
\STATE create a new state $q''$, and set $Q' := Q' \cup {q''}$\COMMENT{add a new state $q''$ to $A'$}
\IF {$q \in Q_+$ or $q' \in Q_+$}
\STATE set $Q_+' := Q_+' \cup \{q''\}$ \COMMENT{$q''$ is accepting if $q$ or $q'$ is accepting}
\ENDIF
\IF {$q \in Q_-$ or $q' \in Q_-$}
\STATE set $Q_-' := Q_-' \cup \{q''\}$ \COMMENT{$q''$ is rejecting if $q$ or $q'$ is rejecting}
\ENDIF
\FORALL[forall transitions with source state $q$ or $q'$] {$t = \left< q_s, q_t, l \right> \in T'$ with $q_s \in \{q,q'\}$}
\STATE $T' := T' \setminus \{\ t \}$ \hfill // remove the transition
\STATE $T' := T' \cup \{\left< q'', q_t, l \right>\}$ \COMMENT{add a new transition with $q''$ as source}
\ENDFOR
\FORALL[forall transitions with target state $q$ or $q'$] {$t = \left< q_s, q_t, l \right> \in T'$ with $q_t \in \{q,q'\}$ }
\STATE $T' := T' \setminus \{\ t \}$ \COMMENT{remove the transition}
\STATE $T' := T' \cup \{\left< q_s, q'', l \right>\}$ \COMMENT{add a new transition with $q''$ as target}
\ENDFOR
\STATE set $Q' := Q' \setminus \{q, q'\}$ \COMMENT{remove $q$ and $q'$ from $A'$}
\WHILE[while non-deterministic choices exist]{ $\left< q_f, q_1, l \right>, \left< q_f, q_2, l \right> \in T'$ with $q_1 \not= q_2$}
\STATE $A'' :=$ {\sf merge}$(A',q_1,q_2)$ \COMMENT{determinize the targets}
\IF{ $A''$ equals {\sc false} }
\STATE \textbf{return} {\sc false}\COMMENT{return {\sc false} if the targets are inconsistent}
\ELSE
\STATE $A' := A''$ \COMMENT{else keep the merge and continue determinizing}
\ENDIF
\ENDWHILE
\STATE \textbf{return} $A'$
\end{algorithmic}
\end{algorithm}
\end{document}